\documentclass[]{bytedance_seed}

\usepackage[toc,page,header]{appendix}

\usepackage{minitoc}

\usepackage{graphicx}
\usepackage{amsmath}

\usepackage{booktabs} 
\usepackage{colortbl}
\usepackage{amsmath}
\usepackage{amssymb}
\usepackage{mathtools}
\usepackage{amsthm}
\usepackage{wrapfig}
\definecolor{skyblue}{RGB}{203, 221, 245}
\usepackage[textsize=tiny]{todonotes}
\usepackage{multirow}
\usepackage{url}
\usepackage{xspace}
\usepackage{graphicx}
\usepackage{microtype}
\usepackage{graphicx}
\usepackage{booktabs} 
\usepackage{colortbl}
\usepackage{amsmath}
\usepackage{amssymb}
\usepackage{mathtools}
\usepackage{amsthm}
\usepackage{adjustbox} 
\usepackage{tcolorbox} 
\definecolor{skyblue}{RGB}{203, 221, 245}
\newcommand{\skyblue}{\rowcolor{skyblue}}
\usepackage{fancyhdr} 
\usepackage[numbers]{natbib}
\usepackage{multicol}

\title{UAM: A Dual-Stream Perspective on \\Forgetting in VLA Training}

\author[*1]{Jianke Zhang}
\author[*2]{Yuanfei Luo}
\author[*1]{Yucheng Hu}
\author[1]{Xiaoyu Chen}
\author[1]{Yanjiang Guo}
\author[2]{Ziyang Liu}
\author[2]{Hongbin Xu}
\author[2]{Tian Lan}
\author[1,\S]{Jianyu Chen}

\affiliation[1]{Tsinghua University}
\affiliation[2]{ByteDance Seed}

\contribution[*]{Equal contribution}
\contribution[\S]{Corresponding Author}

\abstract{
Vision--language--action (VLA) models are typically built by fine-tuning a pretrained vision--language model (VLM) on action data. However, we show that this standard recipe systematically erodes the VLM's multimodal competence, a side effect we call the embodiment tax.
But do VLAs have to forget? Inspired by the two-stream organization of biological vision, we trace this degradation to a structural bottleneck: current VLAs ask a single encoder to support both language-grounded semantics and control-relevant visual features, whereas biological vision separates recognition and visuomotor control into distinct pathways. Building on this view, we propose the Unified Action Model (UAM), which adds a parallel Dorsal Expert, an analog of the brain's dorsal pathway. 
To make the Dorsal Expert an effective second pathway and reduce the control-learning burden on the VLM, we initialize it from a pretrained generative model and train it with a mid-level reasoning objective that predicts visual dynamics.
This design allows us to train the whole VLA end-to-end on action data alone: with no parameter freezing, no gradient stopping, and no auxiliary VL co-training, UAM retains over $95\%$ of the underlying VLM's multimodal capability and at the same time achieves the highest average success rate among baselines on a variety of manipulation tasks that probe out-of-distribution generalization, including unseen objects, novel object--target compositions, and instruction variation. Together, these results suggest that semantic preservation in VLAs can emerge from architectural separation itself, rather than being enforced by frozen weights or auxiliary data replay, and that this preserved semantic capability can naturally transfer from VLMs to semantic generalization in actions.
}

\date{\today}

\checkdata[Email]{\email{\{zhangjk24,chen-xy21,huyc24\}@mails.tsinghua.edu.cn}, \email{luoyuanfei@bytedance.com}}

\checkdata[Project Page]{\url{https://cladernyjorn.github.io/Unified-Action-Model.github.io}}
\begin{document}
\maketitle

\vspace{1ex}
\section{Introduction}
\begin{figure}[ht]
\begin{center}
\includegraphics[width=1.0\linewidth]{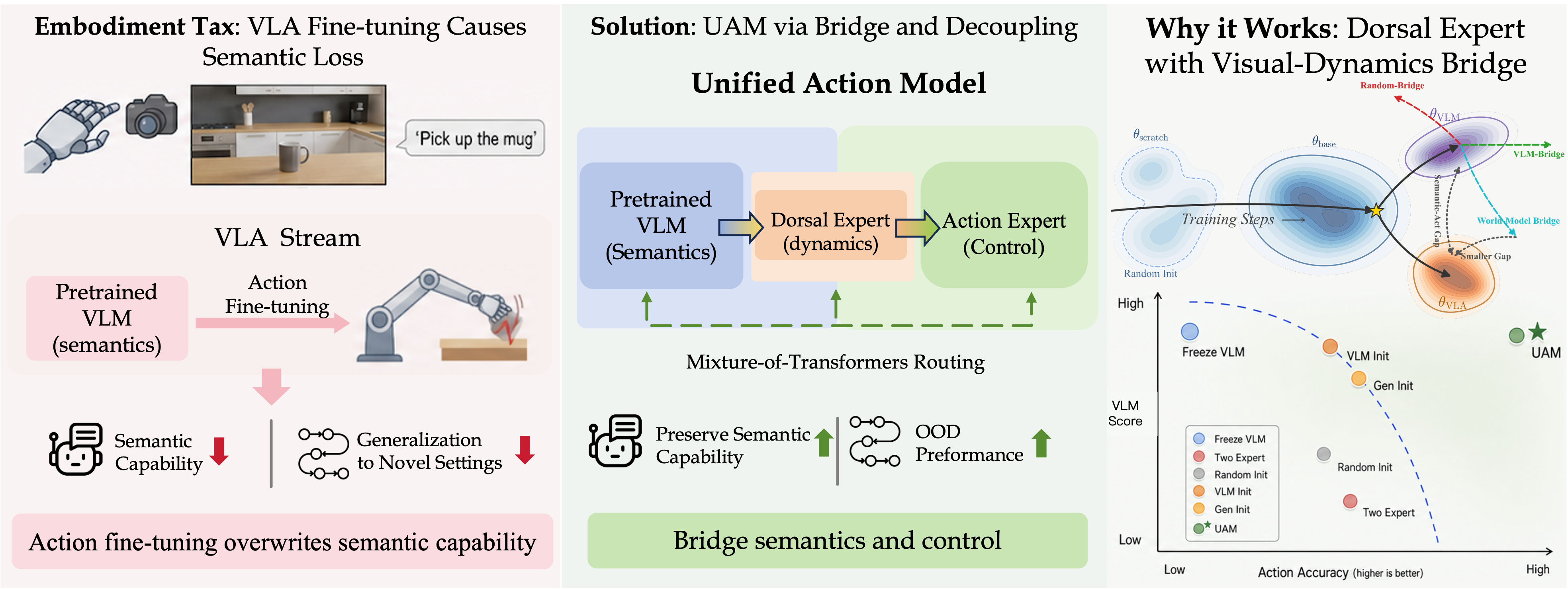}
\end{center}
\caption{\textbf{Reducing the Embodiment Tax with UAM}.
Action-only VLA fine-tuning can erode pretrained VLM capability. UAM separates semantic and control pathways through a visual-dynamics bridge, enabling end-to-end action learning while better retaining multimodal competence.
Direct action fine-tuning overwrites a model's semantic representations, leading to decreased generalization (left). The middle panel introduces the UAM architecture, which utilizes a Mixture-of-Transformers routing to decouple semantics and control via a dorsal expert. The right panel demonstrates that among random initialization, VLM initialization, and generative initialization, we found that the generative expert with visual-dynamics minimizes the semantic-act gap, allowing UAM to achieve high action accuracy without sacrificing VLM performance.
}
\end{figure}
Vision--language--action (VLA) models have rapidly become the default recipe for language-conditioned robot control~\citep{brohan2023rt2,kim2024openvla,octomodelteam2024octo,black2024pi_0,intelligence2504pi05,zhang2025unicod,zhou2025chatvla,chen2025villa}. These systems are usually initialized from a pretrained vision--language model (VLM) and then fine-tuned to emit low-level actions. The motivation is principled: a strong VLM brings hard-won priors, including open-vocabulary recognition, attribute and spatial reasoning, OCR, multilingual grounding, and instruction following, that no robot dataset can reasonably hope to recreate at scale. In this sense, the value of a VLA is not just its action head. Instead, it is the general-purpose multimodal capability that the VLM inherits.

Yet this inheritance is more fragile than is often acknowledged. In Sec.~\ref{sec:method-pre} and Tab.~\ref{tbl:qa_table}, we show across representative VLM backbones and action-head designs that even a moderate amount of action-only fine-tuning is sufficient to measurably erode the pretrained multimodal understanding for which the VLM was chosen.
We refer to this systematic capability loss, in which general-purpose competence is sacrificed as the price of becoming an action model, as the \textit{embodiment tax} of current VLA fine-tuning. 
This degradation matters in practice. The model may still imitate familiar trajectories, but it loses semantic leverage in the long tail, including recognizing unseen objects, handling linguistic variation, and using prior knowledge to disambiguate instructions. The measured regression is therefore not an isolated benchmark artifact. It points to a deeper loss in what the policy can rely on when it must generalize beyond its demonstrations.

Existing remedies mostly preserve the VLM by either withholding or replaying. \textit{Freezing the VLM} protects multimodal competence by construction, but forces control to rely on static features that may not match the spatial and dynamic demands of manipulation~\citep{zhang2026vlm4vla}. \textit{Co-training with auxiliary vision--language data}~\citep{intelligence2504pi05,zhou2025chatvla,lin2026systematic} keeps the gradients flowing, but retention then depends on the scale, diversity, and availability of an auxiliary corpus, often with additional insulation mechanisms~\citep{driess2025knowledge} to prevent objective interference. These methods can be effective, but they sidestep rather than expose the source of the tension: why the \textit{embodiment tax} arises in the first place.

Yet action learning need not, in principle, erase semantic competence: a child can learn to catch a ball without forgetting the word for ``ball'' or that round objects roll. This contrast motivates a sharper question:
\begin{quote}
\textit{Can a VLA retain the general-purpose capability of its underlying VLM \textbf{without} freezing parameters and \textbf{without} relying on auxiliary vision--language data?}
\end{quote}
A positive answer would suggest that semantic preservation can be a property of the architecture itself. Neuroscience offers a direct hint: the \textit{two-streams} organization of primate cortex~\citep{Ungerleider1982TwoCV,goodale1992separate,milner2006visual,rizzolatti2003two,kravitz2011new} separates a \textit{ventral} pathway for object identity and semantics from a \textit{dorsal} pathway for spatial layout and visuomotor control, maintaining functionally distinct representations as motor skills are refined. Current VLAs lack this separation:
a pretrained VLM bears a strong resemblance to the ventral pathway, and fine-tuning it on actions asks the \textit{same} encoder to also support dorsal-like visuomotor functions. 
Because the action loss is denser, lower-level, and trajectory-driven, it overwrites the semantic features that made the VLM valuable. We hypothesize that this {representational bottleneck}, recognition and control forced through a single encoder that nature keeps separate, is a principal driver of VLA forgetting, and the structural root cause that freezing and co-training each treat only symptomatically.

If this hypothesis is correct, the remedy is architectural. We equip the VLA with a Dorsal Expert: a parallel, visually grounded pathway that, by analogy with the dorsal stream, gives the policy a control-oriented visual route in addition to the semantic VLM, without freezing and without auxiliary VL data. The resulting \textbf{Unified Action Model (UAM)} couples a semantic VLM, the Dorsal Expert, and an action expert through parallel mixture-of-transformer layers, so that recognition-oriented and control-oriented visual pathways are present in the architecture rather than collapsed into a single encoder. We then treat the design of the Dorsal Expert as a controlled empirical question (Sec.~\ref{sec:method-prior}), comparing random, VLM-initialized, and generative initializations with or without visual-dynamics supervision~\citep{du2023unipi,hu2024vpp,zhang2025up,zhang2025dreamvla,lv2025f1,li2025uva,liang2025videopolicy,deng2025bagel,hu2026bagelvla}. This study selects the UAM instantiation: a generative Dorsal Expert trained with a visual-dynamics objective, which engages the pathway in mid-level reasoning of its own and thereby unlocks its potential as a genuine dorsal pathway.

The empirical result is that functional separation emerges under fully end-to-end robot training: no parameter is frozen, no gradient is stopped, and no auxiliary VL replay is used.
In this setting, UAM incurs an embodiment tax of less than $5\%$ on average across multimodal benchmarks, retaining over $95\%$ of the underlying VLM's capability (Tab.~\ref{tbl:qa_table}), while visual-routing analysis shows the Dorsal pathway developing action-relevant visual representations (Sec.~\ref{subsec:rep_analysis}). This retained competence carries over to real-world out-of-distribution settings that probe the same VLM capabilities (Sec.~\ref{subsec:eval_manipulation}). UAM gives the best average result among the compared variants, covering unseen objects, new object-target compositions, and instruction variation (Fig.~\ref{tab:real_world_ood}, \ref{fig:app-basicdemo1}, \ref{fig:app-basicdemo2}).

\section{Related Works}

\paragraph{\textbf{Vision-language-action models.}}
Vision-language-action (VLA) models~\citep{black2024pi_0,cheang2024gr2} adapt multimodal foundation models to predict robot actions from visual observations and language instructions. RT-2~\citep{brohan2023rt2} and OpenVLA~\citep{kim2024openvla} represent actions as discrete tokens, while Octo~\citep{octomodelteam2024octo}, \(\pi_0\)~\citep{black2024pi_0}, and \(\pi_{0.5}\)~\citep{intelligence2504pi05} study generalist pretraining, continuous action heads, and heterogeneous robot data. These methods show that VLMs provide useful semantic and instruction-grounding priors, but action fine-tuning can interfere with the pretrained multimodal capabilities. Recent work mitigates this through co-training~\citep{zhou2025chatvla} or knowledge-insulation mechanisms~\citep{driess2025knowledge,yu2026twinbrainvla}, preserving semantics by adding external multimodal signals or restricting which parameters receive control gradients. UAM is complementary: it makes the separation structural, introducing a Dorsal Expert so that a recognition-oriented VLM pathway and a control-oriented visual pathway can emerge through end-to-end robot training.

\paragraph{\textbf{Video models and unified embodied architectures.}}
A separate line of work leverages visual dynamics for robot learning, including text-guided video generation with inverse dynamics~\citep{du2023unipi}, predictive video representations~\citep{hu2024vpp,ye2026dreamzero,bi2025motus}, latent actions~\citep{chen2024igor,bu2025univla}, and recent adaptations of pretrained video models for action generation~\citep{zhang2025dreamvla,pai2025mimicvideo,kim2026cosmospolicy}. Such approaches capture scene evolution and action-conditioned outcomes, yet multimodal understanding and reasoning abilities are not guaranteed when policies are built primarily on video backbones. In parallel, unified understanding-generation models~\citep{team2024chameleon,xie2024show,shi2025lmfusion,deng2025bagel} combine multimodal understanding with visual generation via shared transformers or expert modules, and robotic extensions such as BagelVLA~\citep{hu2026bagelvla} interleave linguistic planning, visual forecasting, and action prediction. UAM shares this unified modeling substrate, but uses it to ask a different question: unlike prior embodied-generation models, which mainly treat visual forecasting as a policy-learning signal, UAM uses visual dynamics to test whether the action-learning burden can be shifted into a separate Dorsal pathway, so that end-to-end robot training improves control without erasing the VLM's general multimodal knowledge.
\section{From Forgetting to a Dorsal Expert}
\label{sec:method}

This section traces a measure--diagnose--design path from the forgetting phenomenon to our architecture. Sec.~\ref{sec:method-pre} formalizes a forgetting metric and uses it to quantify how severely VLA fine-tuning erodes the underlying VLM across representative backbones and couplings.  Sec.~\ref{sec:method-arch} interprets this measurement: it argues that even an MoT-coupled VLA still routes both semantic understanding and control-relevant visual features through a single encoder, and proposes the \textbf{Dorsal Expert} as a second visual pathway that addresses this. Sec.~\ref{sec:method-prior} then turns the question of \textit{what makes a good Dorsal Expert} into a controlled design-space study, and identifies the configuration we adopt as the Unified Action Model (UAM). The main comparison of UAM against external baselines is deferred to Sec.~\ref{sec:experiment}.

\subsection{Setup: Measuring Forgetting Across VLA Architectures}
\label{sec:method-pre}

\paragraph{\textbf{Task formulation.}}
A VLA model $\pi_\theta$ maps an observation $I_{i,t}$ at step $t$ of trajectory $i$ and a natural-language instruction $L_i$ to a low-level action $a_{i,t} = \pi_\theta(I_{i,t}, L_i)$, where $a_{i,t}$ encodes end-effector pose and gripper state. We follow the standard recipe in which $\pi_\theta$ is initialized from a pretrained VLM and fine-tuned on dataset of trajectories $\mathcal{D}_{\text{act}} = \{\tau_i\}_{i=1}^{N}$, where $\tau_i = \{(I_{i,t}, L_i, a_{i,t})\}_{t=1}^{T_i}$.
\paragraph{\textbf{Measuring forgetting.}}
Let $f_{\text{VLM}}$ denote the original (pre-action-tuning) backbone and $f_{\text{VLA}}$ its action-fine-tuned counterpart. We summarize VLM capability by the average score over a standard collection of multimodal-understanding benchmarks (denoted $S(\cdot)$, higher is better; benchmark suite and evaluation protocol in Appendix.~\ref{sec:app-eval}), and quantify forgetting by the relative drop
$\Delta(f_{\text{VLA}}) \;=\; 1 - \frac{S(f_{\text{VLA}})}{S(f_{\text{VLM}})}$,
which equals $0$ when no measurable capability is lost and $1$ when the score collapses to zero under the same evaluation protocol. We will use $S$ and $\Delta$ throughout Sec.~\ref{sec:method}.

\paragraph{\textbf{Empirical setting.}}
To probe how strongly VLA training interferes with VLM features, we attach an action expert to two representative backbones, Qwen2.5-7B~\citep{bai2025qwen2} and PaliGemma~\citep{beyer2024paligemma}, and train the resulting VLA exclusively on action data (no VL co-training). We compare three configurations of how the action expert is coupled to the VLM:
\textit{Freeze-VLM}, the VLM parameters are kept fixed and only the action expert is trained~\citep{zhang2026vlm4vla};
\textit{+MoT}, both the VLM and the action expert are trained, with parallel Mixture-of-Transformers routing as in $\pi_0$~\citep{black2024pi_0};
\textit{+MLP}, both are trained but action prediction is produced by a sequential MLP head appended to the VLM~\citep{zhang2026vlm4vla}.
All three configurations are evaluated on CALVIN for Action Accuracy and on the multimodal benchmark suite for VLM Score $S$. Note that under Freeze-VLM, $S$ coincides with that of the original VLM by construction, providing a top-line reference for $\Delta$. Implementation details are in Appendix.~\ref{sec:app-bgexp}.
\begin{wrapfigure}{r}{0.5\textwidth}
\vspace{0.5ex}
\includegraphics[width=0.5\textwidth]{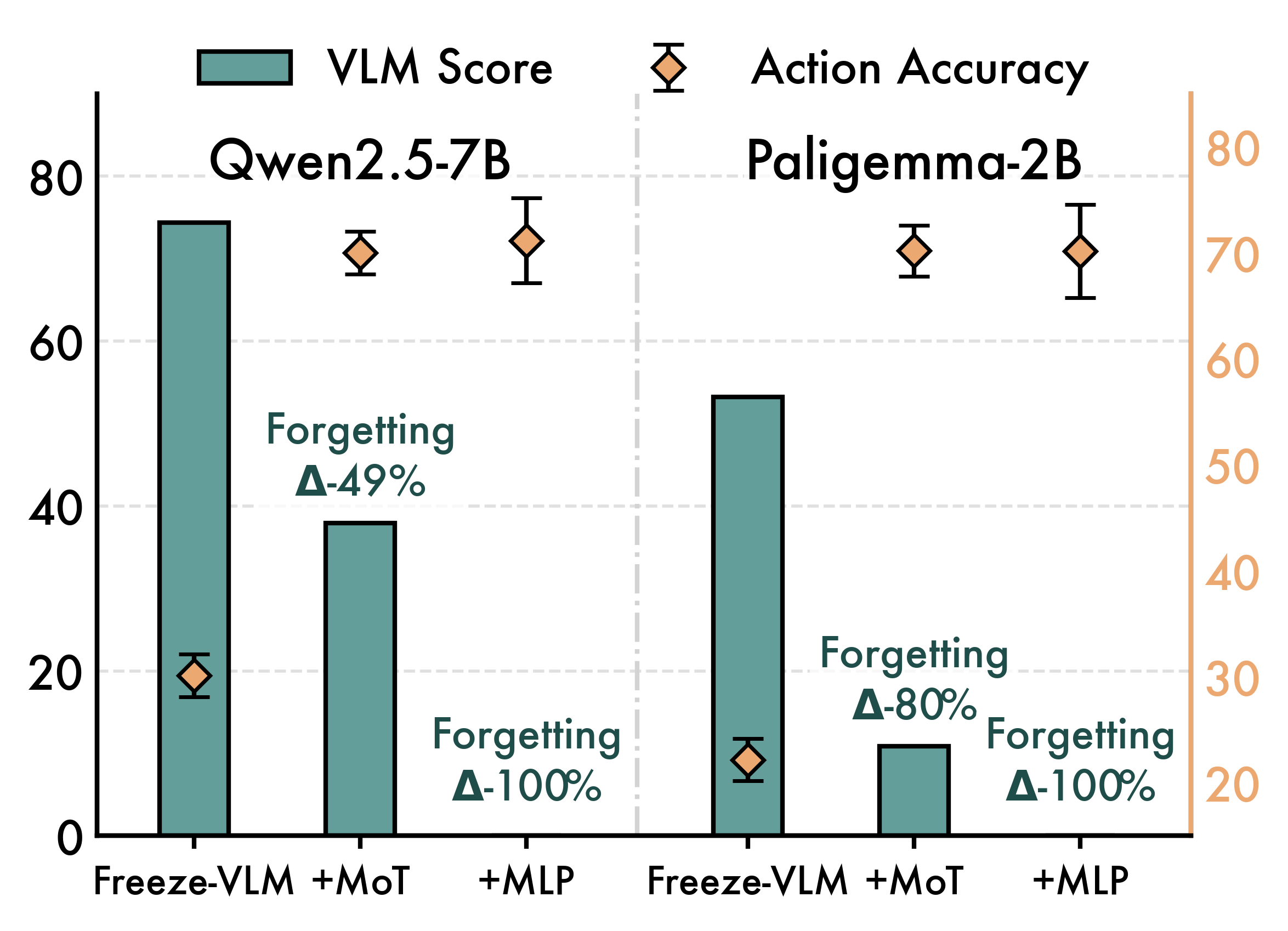}
\vspace{-4ex}
\caption{VLM Score $S$ (green) and Action Accuracy (orange) for VLAs built on two backbones under three couplings: \textit{Freeze-VLM}, \textit{+MoT}, and \textit{+MLP} (Appendix \ref{sec:app-bgexp}).
}
\label{fig:bgexp}
\vspace{-5ex}
\end{wrapfigure}

\paragraph{\textbf{Observations.} }
Fig.~\ref{fig:bgexp} makes the embodiment tax concrete with $\Delta$. Freeze-VLM preserves the VLM Score by construction but yields markedly lower Action Accuracy. Unfreezing the VLM (+MoT or +MLP) raises Action Accuracy sharply, but at a cost: VLM Score drops substantially in both cases, with MoT incurring a clearly smaller drop than MLP at comparable Action Accuracy. The same qualitative pattern holds across both backbones, suggesting that the measured degradation is not tied to a particular VLM. In other words, every unfrozen coupling we tried still pays a sizable embodiment tax. We adopt MoT as the coupling primitive in the rest of Sec.~\ref{sec:method}, since it is the better-performing of the two designs we tested, and ask in Sec.~\ref{sec:method-arch} why even this stronger choice cannot eliminate the tax on its own.

\vspace{3ex}
\subsection{A Bottleneck Hypothesis and the Dorsal Expert}
\label{sec:method-arch}

\paragraph{\textbf{Why does parallel routing not finish the job?}}
A standard VLA, whether MLP- or MoT-coupled, has the form
$a_t = E_{\text{act}}\!\left(E_{\text{sem}}(I_t, L;\theta_{\text{sem}});\,\theta_{\text{act}}\right)$,
in which the backbone $E_{\text{sem}}$ must (i) encode language-grounded semantics for the policy and (ii) supply the visual features through which end-to-end action fine-tuning adapts the model for control. Even when $E_{\text{act}}$ is parameter-isolated via MoT, the visual--linguistic features that $E_{\text{act}}$ consumes still come from $E_{\text{sem}}$, so control-relevant features --- spatial layout, object pose, scene dynamics --- must still be obtained by updating $\theta_{\text{sem}}$. We hypothesize that this is a key source of forgetting: the encoder is a \textit{representational bottleneck} \cite{zhang2026vlm4vla} through which two qualitatively different demands (semantic understanding and visuomotor control) are forced to flow.

\paragraph{\textbf{The Dorsal Expert.}}
Using the dual-stream organization of biological vision~\citep{goodale1992separate} as a design analogy, we propose to give the VLA a \textit{second, dedicated visual pathway} for control-oriented features. We call this pathway the Dorsal Expert $E_{\text{dor}}$. We treat the form of its visual input as part of the design space, with two natural choices: $E_{\text{dor}}$ may consume the raw observation $I_t$ directly, or a set of learnable query tokens $q$. In either case, it produces a set of tokens $Z_{\text{dor}}$ that the action expert may attend to alongside the semantic tokens $Z_{\text{sem}}$ from the original VLM:
\begin{equation}
\notag
    Z_{\text{sem}} = E_{\text{sem}}(I_t, L; \theta_{\text{sem}}),
    \quad
    Z_{\text{dor}} = E_{\text{dor}}(X_{\text{dor}}; \theta_{\text{dor}}),
    \quad
    a_t = E_{\text{act}}(Z_{\text{sem}}, Z_{\text{dor}};\,\theta_{\text{act}}),
\end{equation}
where $X_{\text{dor}}\in\{I_t,\, q\}$. The three experts are coupled by parallel MoT routing. Conceptually, $E_{\text{dor}}$ is intended to reduce the reliance on updating $\theta_{\text{sem}}$ for control-relevant visual features; whether it actually carries this information in practice is an empirical question, addressed quantitatively in Sec.~\ref{sec:method-prior} and qualitatively in Sec.~\ref{subsec:rep_analysis}.

\begin{figure}[t]
    \centering
    \includegraphics[width=1.0\textwidth]{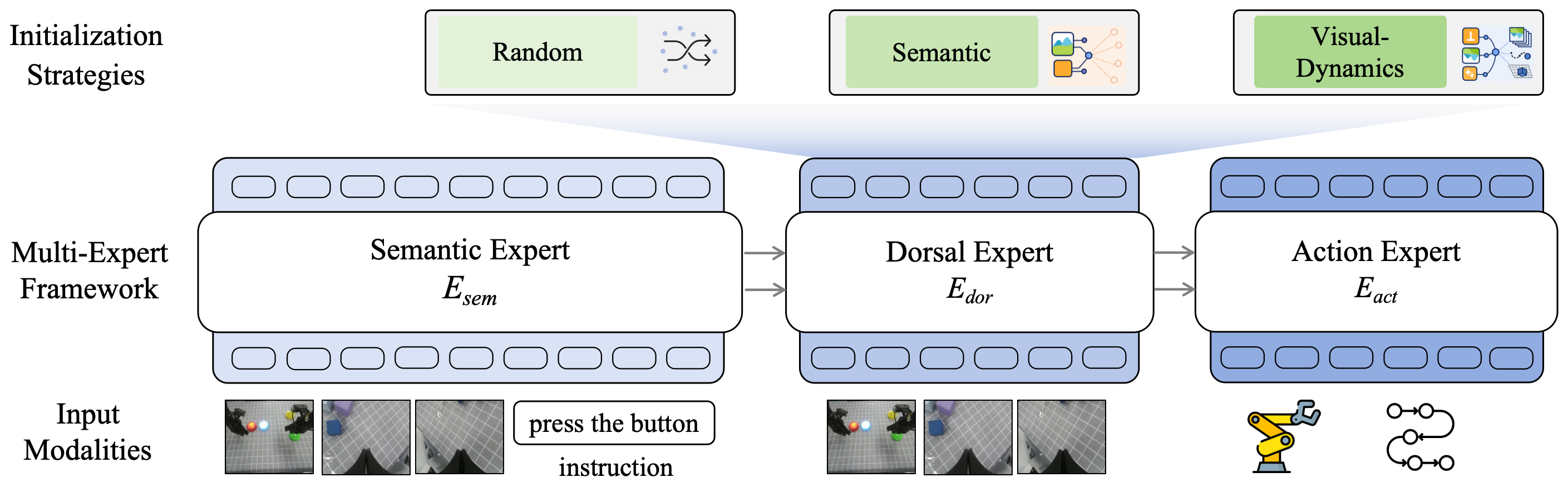}
    \caption{\textbf{Unified Action Model (UAM) and the Dorsal Expert design space.}
    \textbf{(Bottom)} The three-expert macro-architecture: a semantic expert $E_{\text{sem}}$ initialized from a pretrained VLM, a Dorsal Expert $E_{\text{dor}}$, and an action expert $E_{\text{act}}$, coupled via parallel routing.
    \textbf{(Top)} The Dorsal Expert design space we explore in Sec.~\ref{sec:method-prior}, varying input modality and initialization (random, VLM, generative expert with or without an auxiliary visual-dynamics objective). Implementation details are in Appendix.~\ref{sec:app-impl}.}
    \label{fig:uam_architecture}
\end{figure}


\subsection{Designing the Dorsal Expert: A Visual-Dynamics Prior with Matched Supervision}
\label{sec:method-prior}

\paragraph{\textbf{Design space and evaluation protocol.}}
The Dorsal Expert is currently only a hypothesis: Sec.~\ref{sec:method-arch} argues that a second visual pathway should relieve the bottleneck, but does not yet say what such a pathway should actually look like. This subsection turns that question into an empirical study. We take a Dorsal Expert design to be good if it satisfies two criteria: (i) it is at least on par with full fine-tuning on in-distribution action accuracy and improves generalization across our out-distribution real-world manipulation tasks; and (ii) it does not erode the semantic competence of $E_{\text{sem}}$, measured by the forgetting metric $\Delta$ from Sec.~\ref{sec:method-pre}. The two criteria correspond to the two arms of the embodiment tax that motivated the Dorsal Expert in the first place.
We vary the Dorsal Expert along two architectural axes: (a) \textit{initialization}, with three options: random, pretrained VLM, and a pretrained generative unified-multimodal model~\citep{deng2025bagel,hu2026bagelvla}; and (b) \textit{input modality}, either the raw observation $I_t$ or learnable query tokens $q$. All variants share the same MoT routing and the same action expert. We additionally consider an auxiliary visual-dynamics objective $\mathcal{L}_{\text{wm}}$ in one variant in Sec.~\ref{sec:method-para-uam}.  Each variant is trained with $E_{\text{sem}}$ unfrozen, and we additionally evaluate the same trained model with $E_{\text{sem}}$ frozen, using the 2-expert ($\pi_0$-style) baseline as the no-Dorsal reference. Reading the two together separates what each variant achieves on criterion (i) from how much of that achievement actually flows through $E_{\text{dor}}$. We use this six-variant sweep to filter on criterion (i); only the design that survives this filter is then carried forward to the full multimodal evaluation that quantifies criterion (ii).

\begin{table}[h]
\centering
\caption{Design points used to ask, empirically, what a good Dorsal Expert looks like. We sweep two architectural axes (initialization and input modality), and introduce a visual-dynamics objective only for the final variant. The 2-expert ($\pi_0$-style) row is the no-Dorsal reference. The comparison experimental results are shown in Fig~\ref{fig:method-exp}.}
\label{tab:dor_design_space}
\begin{tabular}{lccc}
\toprule
Variant & $E_{\text{dor}}$ initialization & $X_{\text{dor}}$ input & Auxiliary loss \\
\midrule
No Dorsal Expert ($\pi_0$-style) & -- & -- & No \\
Random-init & Random & Visual tokens & No \\
VLM-init input vision (2a) & Pretrained VLM & Visual tokens & No \\
VLM-init input query (2b) & Pretrained VLM & Learnable queries & No \\
Gen-init only (3a) & Generative UMM & Visual tokens & No \\
UAM (3b) & Generative UMM & Visual tokens & Visual-dynamics \\
\bottomrule
\end{tabular}
\vspace{-1ex}
\end{table}
\begin{figure}[h]
\begin{center}
\vspace{-2ex}
\includegraphics[width=1.0\linewidth]{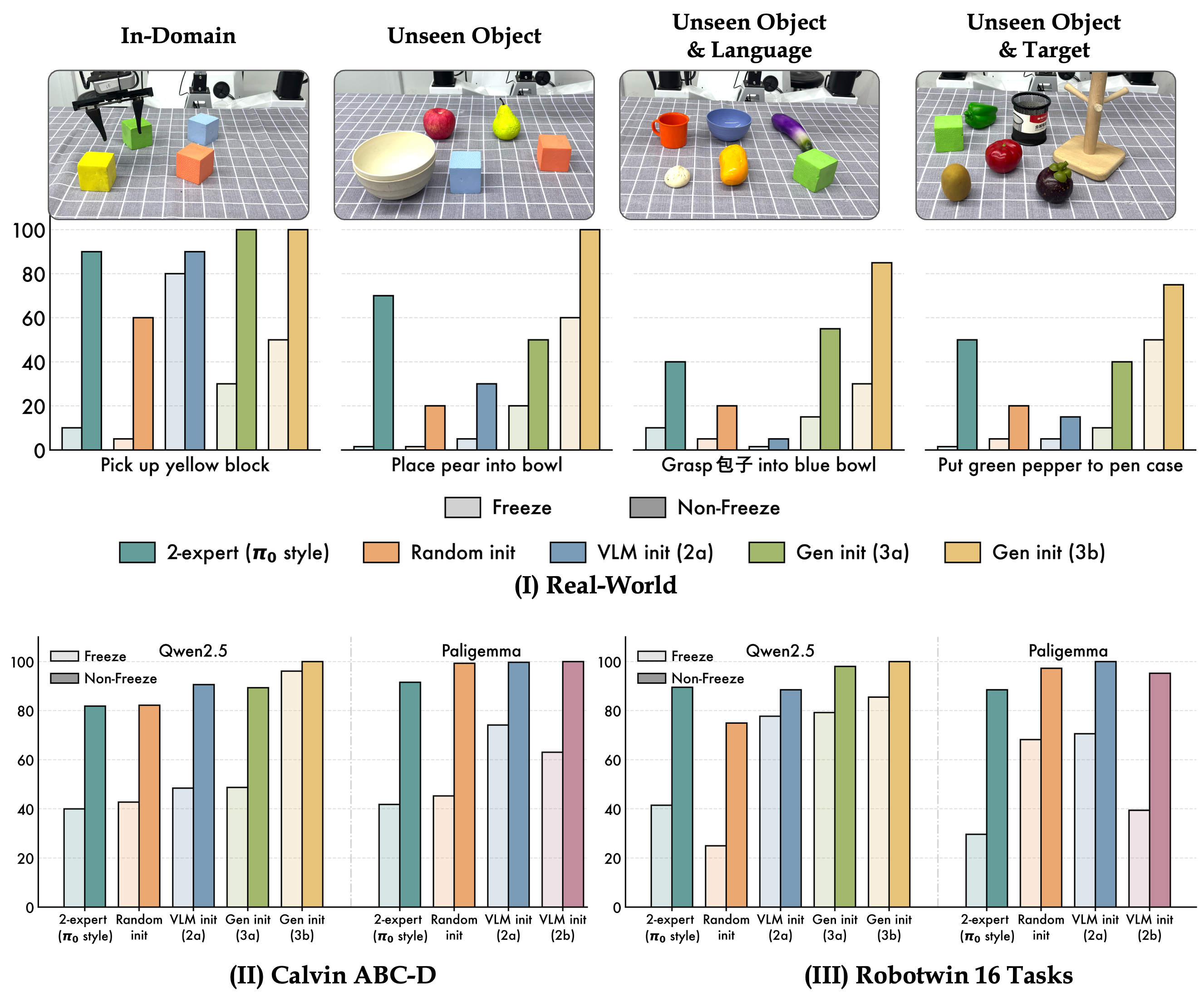}
\vspace{-4ex}
\end{center}
\caption{Dorsal Expert design evaluated on real-world OOD tasks and simulated tasks. For each variant we report results with $E_{\text{sem}}$ frozen and unfrozen. The configuration we adopt as UAM (generative initialization with visual-dynamics supervision) achieves the best trade-off between action performance and robustness when $E_{\text{sem}}$ is frozen. Details are in Appendix.~\ref{sec:app-train}.}
\label{fig:method-exp}
\end{figure}

\paragraph{\textbf{Variant 1: random initialization (no visual prior).}}
A randomly initialized $E_{\text{dor}}$ behaves on par with the 2-expert baseline (Fig.~\ref{fig:method-exp}, ``Random init'') on the simulated benchmarks, but is markedly worse on the real-world tasks, particularly on the OOD splits, and this gap holds whether $E_{\text{sem}}$ is unfrozen or frozen. We read this as evidence that the second pathway, in isolation, fails to acquire the visual structure needed for control: simply adding a same-sized but uninitialized expert does not produce a useful Dorsal pathway, and even hurts when the test distribution drifts away from training. Capacity alone is therefore insufficient, and a Dorsal Expert needs a visual prior that gives the pathway something to start from. We test this by varying the prior in the remaining variants.

\paragraph{\textbf{Variant 2: VLM-based initialization (semantic prior).}}
Initializing $E_{\text{dor}}$ from the pretrained VLM (Fig.~\ref{fig:method-exp}, ``VLM init'') gives the Dorsal pathway a real prior and clearly outperforms Random init across both simulation and real-world tasks. Relative to the 2-expert baseline, the picture is more mixed. \textit{Variant~2a} sits slightly above on the simulated benchmarks but slightly below on the real-world tasks. The Dorsal pathway is no longer empty, but a semantic prior alone does not turn it into a clearly stronger second pathway. The choice of input modality matters here as well. Feeding actual vision tokens (\textit{Variant~2a}) is substantially better than feeding only learnable query tokens (\textit{Variant~2b}) and the gap widens under the frozen probe. This is consistent with the central claim of Sec.~\ref{sec:method-arch}: the Dorsal Expert's value lies in providing an \textit{additional visual pathway}, not in providing more parameters that re-process existing semantic features.

\paragraph{\textbf{Variant 3: generative initialization, with and without visual-dynamics supervision (UAM).}}
\label{sec:method-para-uam}
We finally initialize $E_{\text{dor}}$ from a pretrained generative unified-multimodal model, which carries a prior over visual generation and scene change rather than over recognition and language. We compare two training objectives:
\begin{itemize}
    \item \textit{Variant~3a} (no visual-dynamics loss): action loss only, $\mathcal{L}_{\text{total}}=\mathcal{L}_{\text{act}}$.
    \item \textit{Variant~3b} (with visual-dynamics loss): action loss plus goal-observation prediction:
    \[\mathcal{L}_{\text{total}}=\mathcal{L}_{\text{act}}+\lambda\mathcal{L}_{\text{wm}}(\hat{I}_{t+1}, I_{t+1})\]
\end{itemize}
\textit{Variant~3a} performs similarly to \textit{Variant~2a} across both simulation and real-world tasks, indicating that a generative prior on its own does not bring more benefits than the VLM prior. \textit{Variant~3b} widens this gap substantially: it attains the highest action performance under the unfrozen setting, the smallest gap to that level under the frozen probe, and the strongest results on real-world OOD tasks. Notably, in several scenarios, its performance remains close to the unfrozen 2-expert baseline even when $E_{\text{sem}}$ is frozen, suggesting that the Dorsal pathway carries much of the control-relevant visual information needed by the action expert.

We read these results as two complementary observations on what makes the Dorsal pathway work. First, the modest 3a-over-2a gap, together with the much larger 3a-over-Random gap, says that the kind of prior in $E_{\text{dor}}$ matters: a prior over visual generation and scene structure is a more natural fit for a control-oriented pathway than a recognition-and-language prior, even before any matched objective is added. Second, the jump from 3a to 3b says that having the right prior is not by itself sufficient: the Dorsal pathway only becomes load-bearing when training drives it to perform mid-level reasoning of its own (in our instantiation, visual-dynamics reasoning over how the scene evolves), so that the middle expert is genuinely activated as the second stream. With both pieces in place, the architecture starts behaving like an actual dual-stream system: $E_{\text{sem}}$ continues to carry ventral-style semantics, while $E_{\text{dor}}$ contributes rich control-relevant features on its own, as evidenced by 3b's strong performance under the frozen probe. We therefore adopt \textit{Variant~3b} as the \textbf{Unified Action Model (UAM)}, and report its evaluation against criterion~(ii), whether $E_{\text{sem}}$ is left intact enough to retain the original VLM's multimodal competence, in Sec.~\ref{subsec:eval_multimodal}.

\section{Experiment and Analysis}
\label{sec:experiment}

Our experiments are designed to answer three core questions corresponding to our theoretical claims: 
\textbf{(1)} Does the UAM paradigm effectively mitigate catastrophic forgetting in VLMs during action fine-tuning? (Sec.~\ref{subsec:eval_multimodal})
\textbf{(2)} Does the semantic retention of VLM translate to superior robotic manipulation performance in UAM, especially in out-of-distribution scenarios (unseen tasks during action training)? (Sec.~\ref{subsec:eval_manipulation})
\textbf{(3)} What information is learned by $E_{\text{sem}}$ and $E_{\text{dor}}$ in UAM, respectively? Does UAM automatically route different action-relevant visual signals into these two experts? (Sec.~\ref{subsec:rep_analysis})

\textbf{Implementation Details.} Following the implementation details of UAM (Variant 3b) presented in Sec.~\ref{sec:method-para-uam}, we initialize both the VLM expert and the dorsal expert using the pre-trained Bagel~\cite{deng2025bagel} checkpoint (more details can be found in Appendix.~\ref{sec:app-uam-impl}). Subsequently, the model is trained directly for 30,000 steps on 3k demonstration trajectories collected from the ALOHA bimanual robotic system (without any co-training on additional vision-language data). Under this experimental setting, the subsequent experiments and analyses will evaluate the model's language retention and action semantic generalization, as well as investigate the underlying mechanisms. 

\subsection{Evaluation of Multimodal Understanding}
\label{subsec:eval_multimodal}
\begin{table*}[h]
  \centering
  \caption{Evaluation of MLLMs and VLAs on selected multimodal understanding benchmarks. * denotes VLA methods that use multimodal-understanding/reasoning data for cotraining, while others (OpenVLA and UAM) are trained with action-only data with no parameter frozen.
  }
  \label{tbl:qa_table}
  \resizebox{1.0\linewidth}{!}{
      \begin{tabular}{c|c|cccccccc}
        \toprule
        Method & \#Params & MMMU & MME-P & MME-S & MMBench & MM-Vet & MathVista & MMStar & TextVQA \\
        \midrule
        \skyblue\multicolumn{10}{c}{\textit{Vision Language Models}} \\
        \midrule
        
        Qwen2-VL~\cite{bai2025qwen2}& 1.5B & 41.1&— & 1872 & 74.9 & 49.5 & 43.0 & 48.0 & 79.7 \\
        Qwen2.5-VL~\cite{qwen2.5} & 3B & 53.1 &—& 2157 & 79.1 & 61.8 & 62.3 & 56.3 & 79.3 \\
        LLava-OV~\cite{li2024llava} & 7B & 48.8 &1580& — & 80.8 & 57.5 & 63.2 & 61.7 & — \\
        Qwen2-VL~\cite{bai2025qwen2} & 7B & 54.1 &—& 2327 & 83.0 & 62.0 & 58.2 & 60.7 & 84.3 \\
        Qwen2.5-VL~\cite{qwen2.5} & 7B & 58.6 &—& 2347 & 83.5 & 67.1 & 68.2 & 63.9 & 84.9 \\
        Kimi-VL~\cite{team2025kimi}& 3B/16B & 57.0 &—& — & 83.1 & 66.7 & 68.7 & 61.3 & — \\
        DeepSeek-VL2~\cite{lu2024deepseek} & 4B/27B & 51.1 &—& 2253 & 83.1 & 60.0 & 62.8 & 61.3 & 84.2 \\
        \midrule
        \skyblue\multicolumn{10}{c}{\textit{Unified Multimodal Large Language Models}} \\
        \midrule
        Show-$o_{512}$~\cite{xie2024show} & 1.3B & 26.7 &1097& — & — & — & — & — & — \\
        Janus~\cite{wu2025janus} & 1.5B & 30.5 &1338& — & 69.4 & 34.3 & — & 37.6 & — \\
        Janus-Pro~\cite{chen2025janus}& 7B & 41.0 &1444&  — & 79.2 & 50.0 & — & — & — \\
        LlamaFusion~\cite{shi2025lmfusion} & 8B & 41.7 &1604& — & 72.1 & — & — & — & — \\
        SEED-X~\cite{ge2024seed} & 13B & 35.6 &1457& — & 70.1 & 43.0 & — & — & — \\
        BAGEL~\cite{deng2025bagel} & 7B MoT & 55.3 &1687& 2388 & 85.0 & 67.2 & 73.1 & — & — \\

        \midrule
        \skyblue\multicolumn{10}{c}{\textit{Vision-Language-Action Models}} \\
        \midrule
        OpenVLA~\cite{kim2024openvla} & 7B & 0 & 0 &—& 0 & 0 & 0 & 0 & 0 \\
        ECoT~\cite{zawalski2024robotic}*     & 7B & 5.4 & 0 &—& — & — & — & 0 & 0 \\
        DiVLA~\cite{wen2025diffusionvla}* & 2B & 17.2 & 187 &—& — & — & — & 21.1 & 7.5 \\
        ChatVLA~\cite{zhou2025chatvla}* & 2B & 37.4 & 1435 &—& 69.0 & — & — & 47.2 & 71.2 \\
        $\pi_{0.5}$-base~\cite{intelligence2504pi05}*& 2B MoT & 18.7 & 1032 &1241& 7.3 & — & — & — & — \\
        \midrule
       \textbf{UAM(Ours)} & 7B MoT & \textbf{53.7} &\textbf{1607}& \textbf{2289} & \textbf{83.7} & \textbf{63.4} & \textbf{68.2} & \textbf{61.3} & \textbf{84.2} \\
    \bottomrule
      \end{tabular}
  }
\vspace{-2ex}
\end{table*}

To quantify catastrophic forgetting, we evaluate the semantic retention of various models on multimodal understanding benchmarks, including MMMU~\cite{yue2024mmmu}, MME~\cite{fu2023mme}, MMBench~\cite{yu2023mm}, MathVista~\cite{lu2023mathvista}, MMStar~\cite{chen2024we}, and TextVQA~\cite{singh2019towards}. We compare the various pre-trained VLM (Upper Bound) with VLA models fine-tuned on robotic action datasets.

As demonstrated in Tab~\ref{tbl:qa_table}, we observe that when subjected to full-parameter fine-tuning (encompassing the VLM, Dorsal Expert, and Action Expert) and trained to converge exclusively on action-only data, the UAM preserves its inherent vision-language capabilities. Its performance remains comparable to state-of-the-art MLLMs across multiple VLM benchmarks. Results of OpenVLA~\cite{kim2024openvla} show that standard end-to-end VLA training without VQA co-training leads to a complete loss of the VLM's multimodal understanding capabilities. Meanwhile, extensive VQA and action data co-training (as seen in methods like ChatVLA~\cite{zhou2025chatvla}, $\pi_{0.5}$ with KI~\cite{intelligence2504pi05,driess2025knowledge}) still compromises the general understanding capabilities of VLM due to the inherent limitations of co-training datasets and the persistent gap between action and QA objectives. In contrast, the proposed UAM automatically preserves the capabilities of its original VLM backbone (with only a marginal performance drop of less than 5\%), without incorporating VQA co-training or relying on gradient-blocking techniques. 


\subsection{Evaluating Robotic Manipulation}
\label{subsec:eval_manipulation}
In this section, we show that alongside the remarkable retention of multimodal understanding, UAM also exhibits robust semantic generalization in action tasks.
We deploy UAM on the bimanual robot platform for complex, unseen manipulation tasks, specifically focusing on semantic generalization (e.g., novel objects, unseen distractions, and complex language instructions). Specifically, after training the UAM on limited (3k) expert trajectories, we deploy it on a real-world robotic platform for evaluation. We compare its performance against Qwen-$\pi_0$ (2-expert) and Variant 2a (3-expert architecture in which $E_{\text{dor}}$ is initialized with a pretrained VLM). More details are in Appendix.~\ref{sec:app-ood}.

\begin{figure}[h]
\begin{center}
\vspace{-1ex}
\includegraphics[width=1.0\linewidth]{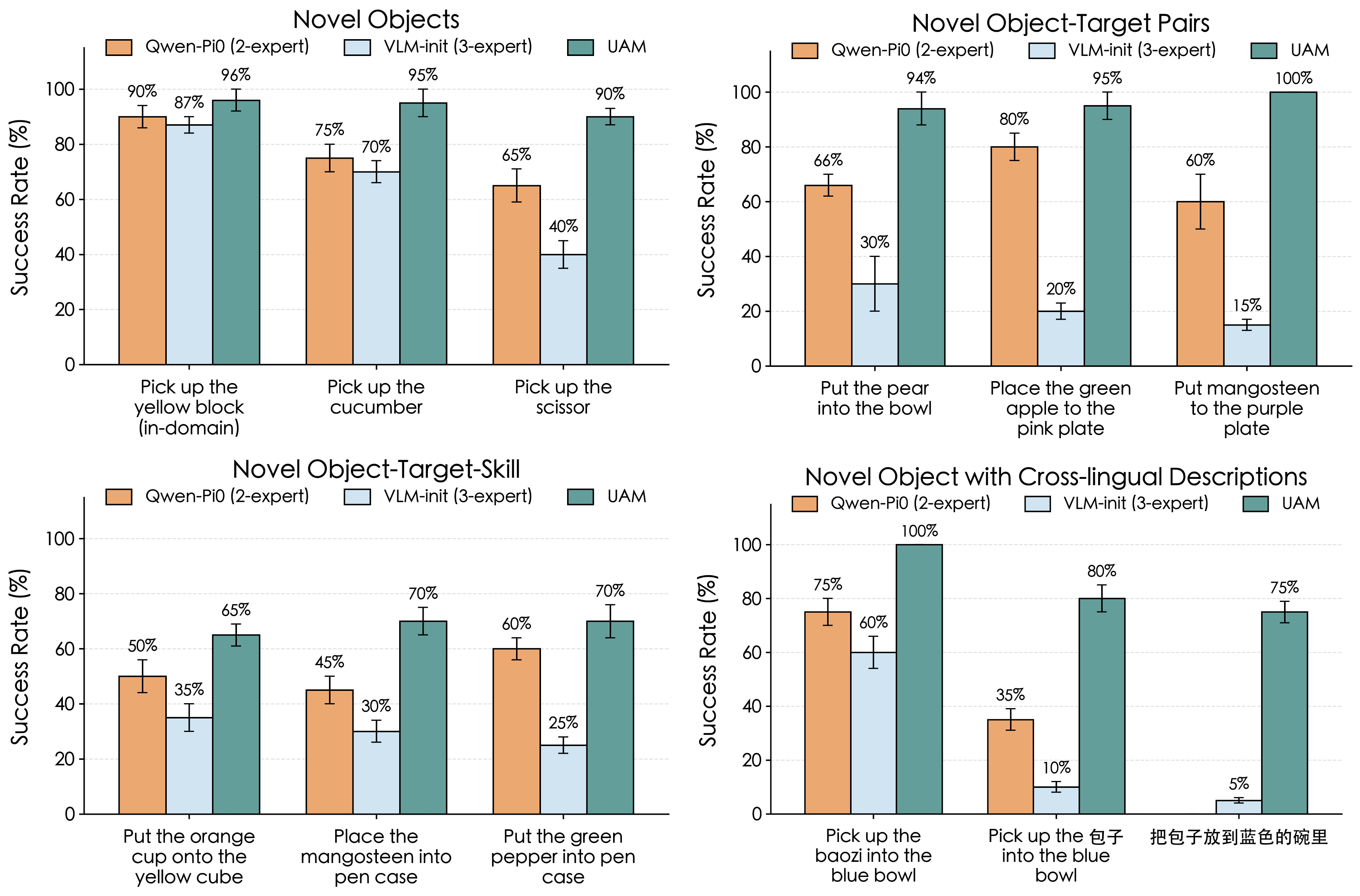}
\vspace{-2ex}
\end{center}
\caption{Success rates (\%) of different models in out-of-distribution (OOD) manipulation tasks. The last sub-chart probe instruction-following ability under language variation, including pinyin transliterations (e.g., ``baozi''), Chinese-English code-mixing, and fully Chinese instructions. More visualization of the OOD settings can be found in Fig.~\ref{fig:app-basicdemo1},~\ref{fig:app-basicdemo2}.}\label{tab:real_world_ood}
\vspace{-2ex}
\end{figure}

As shown in Fig.~\ref{tab:real_world_ood}, we demonstrate that UAM can achieve remarkable semantic generalization using only a limited amount of action-annotated data. This further substantiates that preserving the inherent semantic capabilities of the VLM significantly enhances the sample efficiency of action-annotated data, enabling strong generalization with minimal action data. 

Notably, we observe that the model with a VLM-initialized dorsal expert (VLM-init) achieves comparable performance to the 2-expert baseline on in-domain tasks (90\% vs. 87\%). However, it underperforms the 2-expert model (which lacks a dorsal expert) across all other out-of-distribution (OOD) tasks. This indicates that merely initializing the intermediate dorsal expert with VLM weights fails to effectively transfer the semantic capabilities preserved in $E_{sem}$ to action generalization. This limitation likely stems from the fact that the dorsal expert receives only visual signals as input; consequently, its encoded representations are purely visual, preventing the action policy from attending to the correct linguistic context. In contrast, while the dorsal expert in UAM also exclusively processes visual signals, it exhibits significantly superior action generalization capabilities. This demonstrates that employing a World Model as the dorsal expert effectively bridges the inherent gap between vision-language representations and action execution.

\subsection{Representational Analysis of Visual Signal Routing}
\label{subsec:rep_analysis}

To qualitatively validate our hypothesis that the UAM architecture automatically routes and decouples visual information, we visualize the attention maps during action generation. Specifically, we examine how the action head attends to visual tokens originating from the Semantic Expert ($E_{\text{sem}}$, encoded via ViT) versus the tokens from the Dorsal Expert ($E_{\text{dor}}$, encoded via VAE). 

\begin{figure}[h]
\begin{center}
\includegraphics[width=0.9\linewidth]{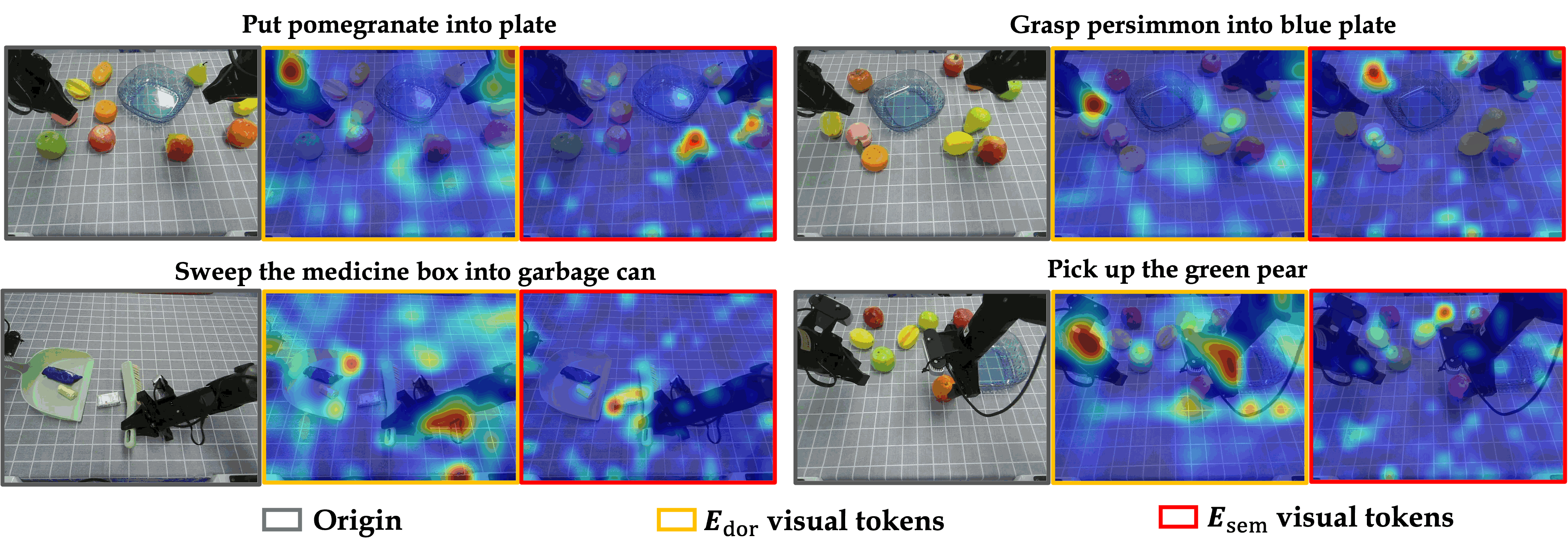}
\end{center}
\caption{Visualization of attention maps during action generation (Appendix.~\ref{sec:app-rep}). For each task, we show the input third-view image, attention over $E_{\text{dor}}$ visual tokens, and $E_{\text{sem}}$ visual tokens. The action queries attend to different regions: attention over $E_{\text{sem}}$ tokens is concentrated on task-relevant semantic entities, such as target objects and goal regions, while attention over $E_{\text{dor}}$ tokens focuses more on the robot arm, interaction regions, and global scene context.}\label{fig:exp-repre}
\end{figure}

As illustrated in Fig.~\ref{fig:exp-repre}, the attention on the $E_{\text{sem}}$ tokens is highly concentrated on the semantic elements of the scene, 
which confirms that the bridge effectively shields $E_{\text{sem}}$, allowing it to retain its original VLM capabilities and act as the ``What'' stream for robust semantics alignment. Conversely, the attention distribution on the $E_{\text{dor}}$ tokens exhibits a distinctly different pattern, shifting heavily toward the current state of the robotic arm and broader global contextual information. This observation suggests that the intermediate world modeling module forces $E_{\text{dor}}$ to capture the physical state and visual evolution of the scene, effectively functioning as the ``Where/How'' stream. Overall, these visualizations provide compelling evidence that UAM naturally induces a functional bifurcation. By automatically routing the visual demands of action signals into semantic-centric and dynamic-centric pathways, we elegantly resolve the representational conflict inherent in standard VLA models.
\section{Conclusion}
We studied the fragility of VLM inheritance in VLA fine-tuning. Our results show that action-only adaptation can measurably reduce the pretrained multimodal capability that makes VLMs useful for robot learning, an effect we call the embodiment tax. We further examined whether this loss is related to forcing semantic recognition and visuomotor control through the same pathway. In a three-expert VLA formulation, we found that simply adding a bridge or initializing the bridge from a VLM is not enough. Instead, visual-dynamics initialization and supervision are important for shaping the bridge into an action-relevant visual route.

This study leads to UAM, an end-to-end framework with a generative Dorsal Expert and a visual-dynamics objective, using no frozen parameters, stopped gradients, or auxiliary VL replay. UAM retains over \(95\%\) of the VLM's semantic capability on average and performs best among compared variants in the OOD settings. However, the world-model-based bridge introduces additional training and inference complexity, which can be a potential limitation. Overall, we view UAM as a flexible framework for studying embodied transfer and a foundation for future work on richer semantic, predictive, and control-aware interfaces.
\section{Acknowledgements}
We sincerely thank Sheng Chen and Yangang Zhang for their strong support and fruitful discussions.



\bibliographystyle{plainnat}
\bibliography{main}

\clearpage

\beginappendix

\section{Evaluation Details in Methodology}
\label{sec:app-eval}
\paragraph{Evaluation Protocol} 
To systematically investigate the forgetting phenomenon and the effectiveness of our proposed solution, we decouple the evaluation of VLA models into two distinct dimensions:
\begin{itemize}
    \item \textit{Semantic Retention (Forgetting Metric):} We evaluate the fine-tuned VLA models on standard multimodal understanding benchmarks (MMMU, MME, and MMBench) to quantify the preservation of the VLM's pre-trained general knowledge and reasoning abilities.
    \item \textit{Action Performance (Control Metric):} To ensure that semantic retention does not compromise control accuracy, we rigorously test the action success rate in both simulation and the real world. 
    \begin{itemize}
    \item For simulation, we utilize the Calvin (ABC-D split)~\citep{mees2022Calvin} and RoboTwin~\citep{chen2025robotwin} for the in-domain manipulation test. 
    \item For the RoboTwin benchmark, to mitigate the prohibitive time costs associated with evaluation, we curated a test suite comprising 16 tasks. Specifically, from the original 50 tasks, we selected 1 to 2 tasks per skill category that exhibited the lowest success rates, thereby representing the most challenging scenarios. The models were trained and subsequently evaluated on this suite (utilizing 50 expert demonstrations per task), which includes the following tasks: Beat Block Hammer, Blocks Ranking Rgb, Blocks Ranking Size, Click Alarmclock, Dump Bin Bigbin, Handover Block, Hanging Mug, Lift Pot, Move Can Pot, Pick Diverse Bottles, Place Can Basket, Place Dual Shoes, Put Bottles Dustbin, Scan Object, Stack Blocks Three, Stamp Seal, and Turn Switch.
    \item
        For real-world deployment, we evaluate the models on a dual-arm ALOHA system, assessing out-of-distribution (OOD) semantic generalization capabilities (unseen objects during action-only finetuning). Following training on the 3,000 collected demonstration trajectories (without any prior pre-training), we evaluate the model across various OOD tasks, which target objects that are unseen in the training set, alongside the deliberate introduction of additional distractor objects. Detailed tasks are shown in Fig~\ref{fig:method-exp} and Appendix~\ref{sec:app-ood}.
    \end{itemize}
\end{itemize}

This dual-metric evaluation framework is designed to simultaneously verify two crucial aspects: first, whether the VLA method retains its semantic understanding alongside strong manipulation proficiency after training exclusively on action data; and second, whether it can effectively transfer its inherent vision-language generalization capabilities to the manipulation of out-of-distribution (OOD) objects in real-world robotic deployments.

\section{Empirical Observations on the Forgetting Dilemma}
\label{sec:app-bgexp}

We initialize the Qwen2.5-7B model using VLM parameters from Bagel~\citep{deng2025bagel} (which empirically outperforms the original Qwen2.5VL-7B) and the PaliGemma model using PaliGemma-1. Building upon these two VLMs, we construct VLAs by integrating action experts in two distinct ways: a parallel Mixture-of-Transformers (MoT) architecture following $\pi_0$~\cite{black2024pi_0}, and a sequential MLP action head following VLM4VLA~\cite{zhang2026vlm4vla}. We perform full-parameter fine-tuning on these models using the Calvin ABC dataset for 10k steps. Specifically, we employ a learning rate of $1 \times 10^{-5}$ for the Qwen-7B series and $5 \times 10^{-5}$ for the PaliGemma series.
Subsequently, we evaluate the understanding capabilities of the VLM component and the overall action performance of the VLA. Note that the simulation evaluation here primarily assesses in-domain action accuracy rather than generalization to novel objects.

The results, presented in Tab~\ref{tab:bgexp} and Fig~\ref{fig:bgexp}, reveal that even when the action policies have largely converged, the degree of retained VLM performance varies significantly across different architectures. For the baseline where the VLM is frozen, although it entirely preserves the model's intrinsic capabilities and yields the highest understanding scores, it suffers from severely degraded action performance. Conversely, all approaches that unfreeze the VLM incur a degradation in its original capabilities. Specifically, the VLA employing the 7B-LLM and the MoT architecture exhibits the best retention of VLM capabilities. In contrast, while the sequential MLP head achieves comparable action accuracy to the MoT architecture, its inherent VLM capabilities are catastrophically destroyed (yielding a VQA score of 0).

\begin{table}[h]
\caption{Detailed results of Fig~\ref{fig:bgexp}. 
For the ``Action'' metric, we report the normalized average completion rate on the test simulation environment as the indicator of action performance. For the ``VLM'' capability, we use the normalized average score across three standard VQA benchmarks as the evaluation metric.
}
\label{tab:bgexp}
\vspace{2mm}
\centering
\begin{adjustbox}{width=0.9\textwidth}
\begin{tabular}{l | c | c | c | c | c}
\toprule
\textbf{VLA Arch (VLM+Action Head)} & \textbf{Action} & \textbf{VQA-AVG} & \textbf{MME} & \textbf{MMMU} & \textbf{MMBench} \\
\midrule
Qwen2.5-Freeze + MLP (7B+10M) & 30.12 & 74.32 & 2354 & 53.7 & 85.2 \\
Qwen2.5 + MoT (7B+1B) & 65.98  & 37.94 & 1675 & 30.46 & 23.54 \\
Qwen2.5 + MLP (7B+10M) & 71.14  & 0 & 0 & 0 & 0 \\
\midrule
Paligemma-Freeze + MLP (2.3B+10M) & 22.18 & 53.18 & 1670 & 34.66 & 65.23 \\
Paligemma + MoT (2.3B+0.3B) & 70.18  & 10.83 & 400 & 6.2 & 12.00 \\
Paligemma + MLP (2.3B+10M) & 70.12  & 0 & 0 & 0 & 0 \\
\bottomrule
\end{tabular}
\end{adjustbox}
\end{table}

\paragraph{\textbf{A more in-depth Analysis}}
When subjected to direct end-to-end action fine-tuning, both models exhibited a severe degradation in VQA performance, confirming the catastrophic forgetting dilemma. The results also yield two critical empirical insights regarding visual-language retention:
\begin{enumerate}
    \item \textit{Parallel Architectures Mitigate Forgetting:} We observe that although action accuracy remains comparable regardless of the architecture used, models employing a MoT architecture retain their language capabilities better than a standard sequential action head. This implies that routing visual and linguistic tokens through decoupled pathways reduces modality interference.
    \item \textit{Model Scale Correlates with Retention:} Model capacity plays a pivotal role. Larger foundational VLMs (7B) demonstrate higher resilience to catastrophic forgetting compared to smaller models (2B), maintaining a higher relative percentage of their original semantic reasoning scores.
\end{enumerate}
While scaling up model size or employing MoT can partially alleviate the symptoms of forgetting, they do not fundamentally resolve the underlying representational conflict between semantic understanding and low-level control.

\section{Dorsal Expert Design Implementation Details}
\label{sec:app-impl}
In this section, we provide detailed implementations of the ablation variants for the Dorsal Expert discussed in Section~\ref{sec:method-arch}. Specifically, we integrate two or three experts using the Mixture-of-Transformers (MoT) mechanism, enabling information exchange among the multiple experts through the design of distinct attention masks.
\begin{figure}[h]
\begin{center}
\vspace{-1ex}
\includegraphics[width=1.0\linewidth]{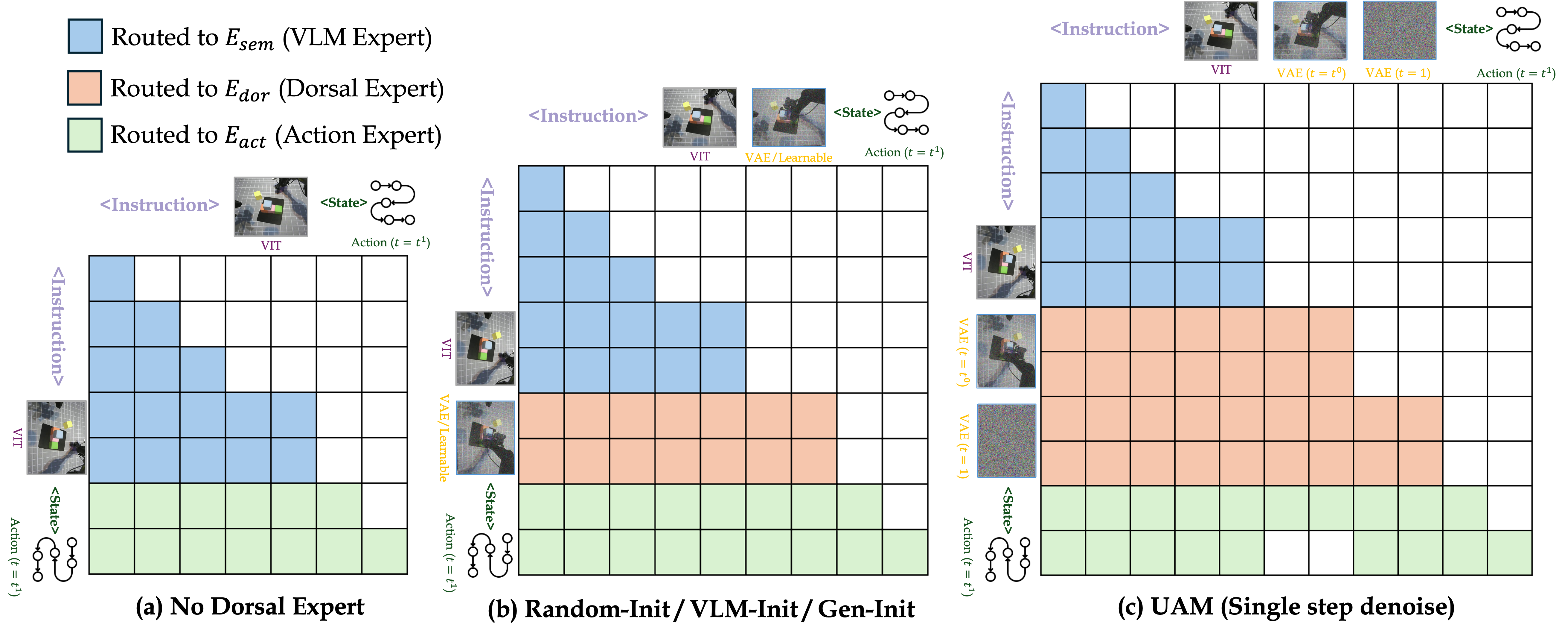}
\vspace{-4ex}
\end{center}
\caption{Attention Mask used for Different Dorsal Expert Design. We visualize the input sequence formats and their corresponding attention masks across different architectural designs (where white regions indicate masked attention). Additionally, distinct colors are employed to denote how different types of tokens are routed to their respective experts by the MoT mechanism.
}\label{fig:app-attn}
\vspace{-2ex}
\end{figure}

\subsection{No Dorsal Expert (2-Expert)}
\label{sec:app-pi0-impl}
We establish the architecture without the Dorsal expert as our baseline. Specifically, following the approach of $\pi_0$~\cite{black2024pi_0}, we couple the VLM expert and the action expert via the MoT, while directly feeding the proprioceptive state into the action expert. The detailed attention implementation is illustrated in Fig~\ref{fig:app-attn}(a).

\subsection{Random-init, VLM-init and Gen-init}
\label{sec:app-vlm-impl}
The implementations of the Random-init, VLM-init, and Gen-init variants share a similar architectural design, differing solely in the initialization weights of the intermediate Dorsal Expert. Specifically, for the Random-init variant, the Dorsal Expert is initialized using standard random weights. For VLM-init, we directly duplicate the weights of $E_{sem}$ (the VLM expert) to initialize the Dorsal Expert. For Gen-init, it is initialized using the pre-trained weights of the Gen expert from the pretrained UMM model (Bagel~\cite{deng2025bagel}). All three variants employ a similar attention masking strategy, as depicted in Fig~\ref{fig:app-attn}(b).

\subsection{UAM}
\label{sec:app-uam-impl}
The implementation of UAM builds upon the Gen-init variant by incorporating an additional world modeling supervision signal. We observe that the prior alignment between the dynamics expert and the VLM is crucial. Consequently, we implement this approach exclusively on the Qwen model, utilizing the pre-trained Bagel~\cite{deng2025bagel} weights (as applying it to the PaliGemma series would necessitate extensive additional alignment between multimodal understanding and generation). Specifically, following the implementation of BagelVLA~\cite{hu2026bagelvla}, we simplify its original image input format and discard the large-scale embodied pre-training phase, opting instead to train the model directly on action data. Fig~\ref{fig:app-attn}(c) illustrates the attention masking strategy and the input sequence concatenation scheme for UAM. Notably, tokens encoded by the ViT are routed to $E_{sem}$ (the VLM expert), whereas tokens encoded by the VAE are directed to $E_{dor}$ (the Dorsal expert). Finally, we adopt the single-step denoising mechanism from BagelVLA to facilitate the interaction within the dual flow-matching framework.

\section{Details and Demos of Real World OOD Tasks}
\label{sec:app-ood}

On the real-world bimanual robotic platform, we train the UAM using 3,000 collected demonstration trajectories. Specifically, the tasks within this training dataset can be broadly categorized into the following types:

\begin{itemize}
    \item \textbf{Pick \& Place:} Grasping and placing objects. The training set includes toy fruits, a computer mouse, colorful blocks, toy phones, and so on. The placed targets include colorful plates, baskets, boxes, and so on.
    \item \textbf{Water Flower:} grasping the handle of a toy watering can to simulate the pouring action of watering a plant.
    \item \textbf{Stack Cubes:} Stack blocks of four different colors.
    \item \textbf{Stack Bowls:} Stack bowls of three different colors according to a specified color sequence.
    \item \textbf{Pour Fries:} Open the lid of a carton and pour the toy fries from inside it onto a plate.
    \item \textbf{Sweep Rubbish:} Grasp a toy broom, sweep the randomly placed tissue paper trash on the table into a dustpan, and then put down the broom.
    \item \textbf{Press Button:} Press different buttons in a specified color sequence.
    \item \textbf{Drawer Operation:} Opening and closing a drawer.
    \item \textbf{Other Complex Motion Tasks:} Additionally, the training dataset further includes tasks that require complex manipulations, which encompass: wiping a table with a cloth, inserting flowers into a vase, hanging a cloth, and plugging in a power cable, and so on. These tasks primarily demand high manipulation precision, placing minimal requirements on the model's semantic generalization capabilities.

\end{itemize}

\begin{figure*}[h]
    \centering
    \includegraphics[width=\textwidth]{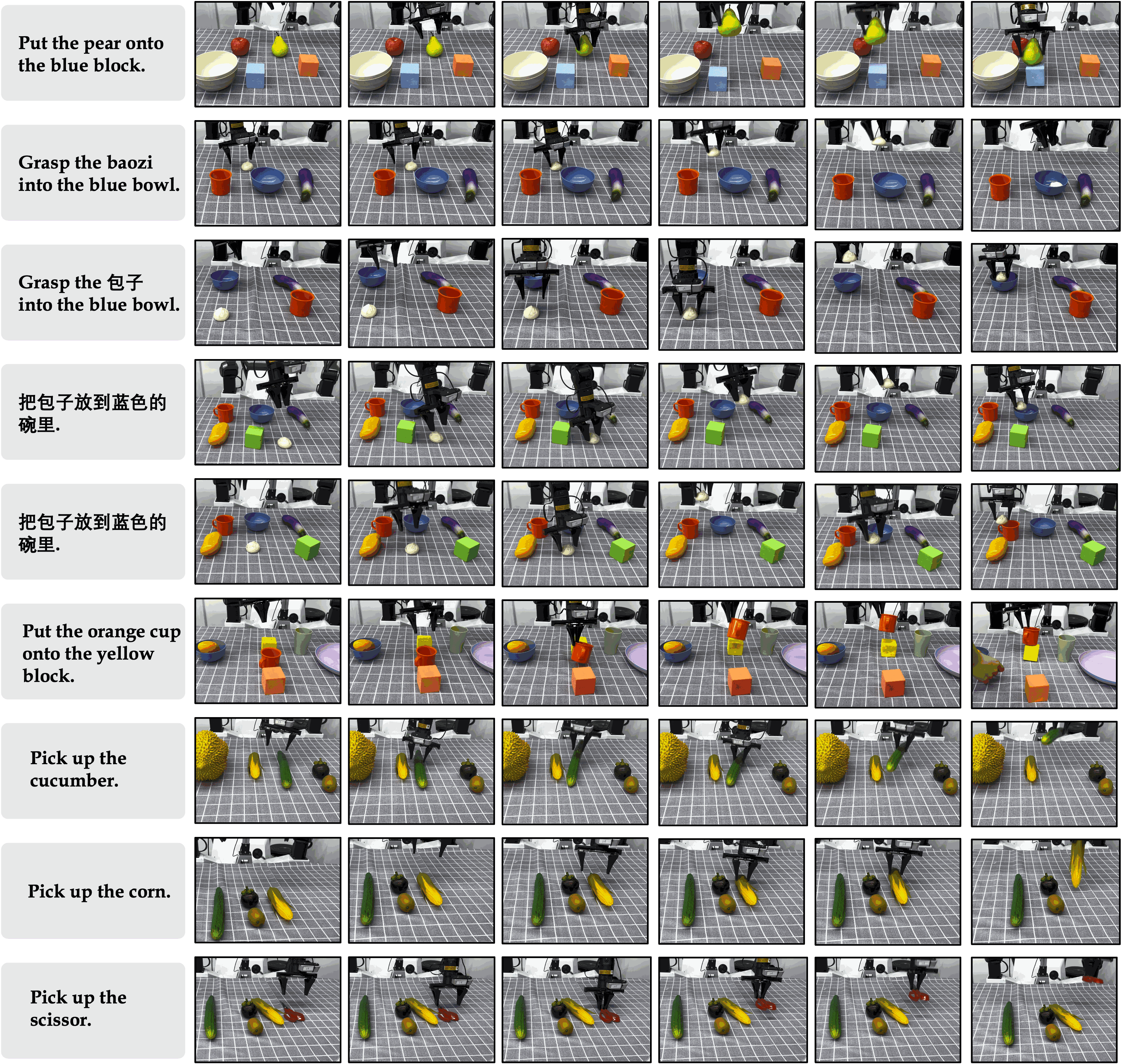}
    \caption{\textbf{Evaluation videos of UAM on semantic out-of-distribution tasks}. The text on the left denotes the input language instructions. All demonstrated scenarios are unseen during training.
}
    \label{fig:app-basicdemo1}
\end{figure*}

\begin{figure*}[h]
    \centering
    \includegraphics[width=\textwidth]{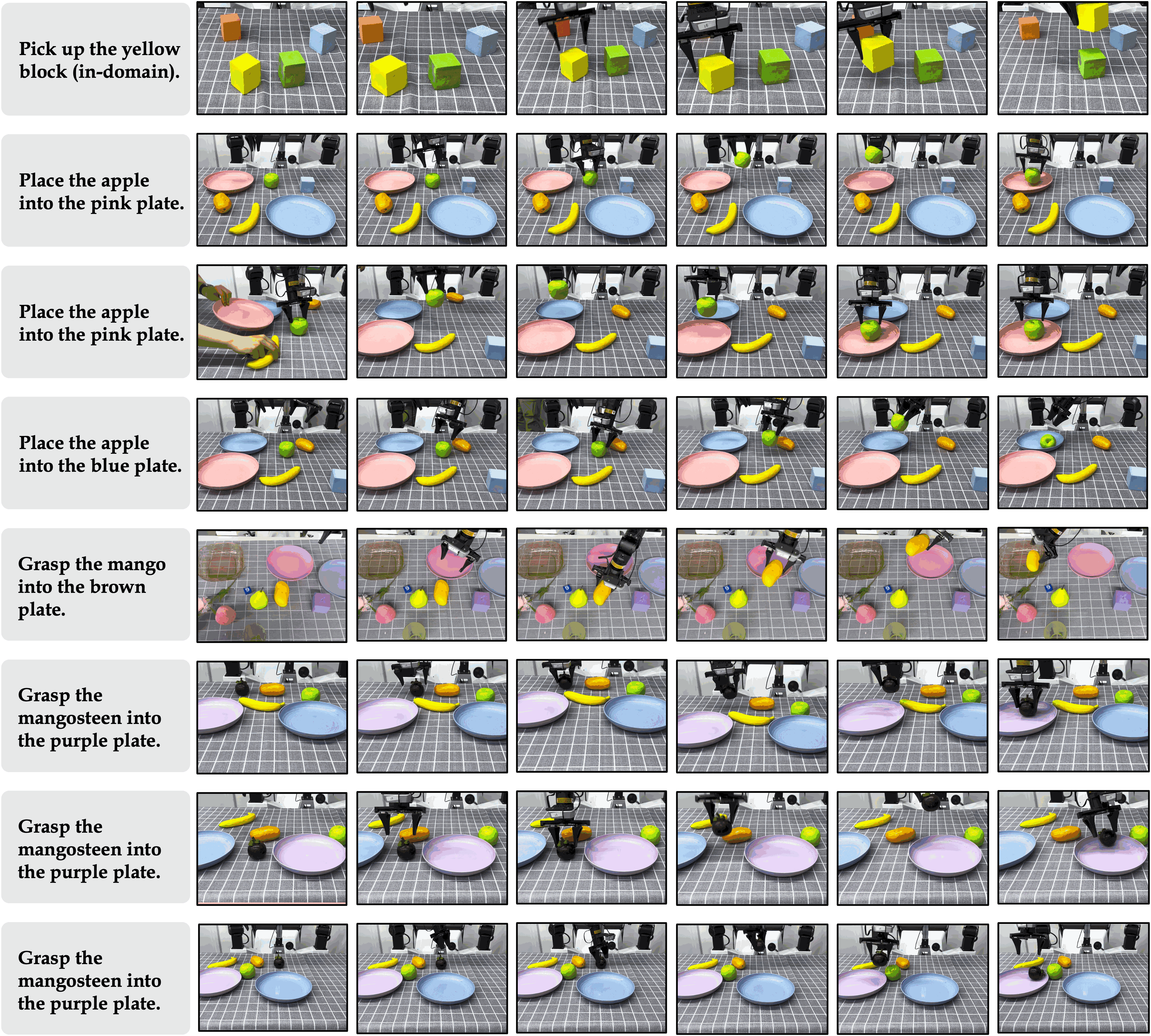}
    \caption{\textbf{Evaluation videos of UAM on semantic out-of-distribution tasks}. The text on the left denotes the input language instructions. All demonstrated scenarios are unseen during training except ''in-domain".
}
    \label{fig:app-basicdemo2}
\end{figure*}

During evaluation, our primary objective is to verify whether the model possesses sufficient semantic generalization capabilities. Specifically, we utilize pick-and-place, placement, and recognition tasks to assess the model's ability to manipulate elements unseen during training, including novel objects, novel targets, novel distractor objects, and novel language instructions (or linguistic variations). We evaluate the model across a diverse range of such out-of-distribution (OOD) tasks, including the real-world evaluation scenarios depicted in Fig~\ref{fig:method-exp} and \ref{tab:real_world_ood}. Demos of evaluation demonstrations are also provided in Fig~\ref{fig:app-basicdemo1} and Fig~\ref{fig:app-basicdemo2}.


\section{Implementation Details of Attention Visualization}
\label{sec:app-rep}

In this section, we provide the core implementation details for the attention heatmaps presented in Figure 9 and further analyze the emergent attention decoupling phenomenon observed in our dual-stream architecture.
\subsection{Visualization Pipeline}
To visualize the attention during action generation, we extract the MoT self-attention weights from the action prediction tokens (Queries) to the visual tokens (Keys) in the middle of self-attention layers (14/28). The weights are averaged across all attention heads and action steps to obtain a stable, global distribution.
The core challenge of visualization lies in accurately reconstructing the 2D spatial grids from the flattened 1D visual tokens, as the two visual encoders employ different downsampling strategies. To preserve the original $640 \times 480$ (4:3) aspect ratio and avoid spatial misalignment caused by forced square cropping, the preprocessing module dynamically adjusts the input resolution:
\begin{itemize}
    \item \textbf{Semantic Visual Tokens} ($E_{sem}$): Based on a ViT architecture with a patch size of 14, it generates 252 visual tokens, which are reshaped into a $14 \times 18$ (Height $\times$ Width) 2D grid.
    \item \textbf{Dorsal Visual Tokens} ($E_{dor}$): Based on a VAE architecture with a downsampling stride of 16, it generates 192 visual tokens, which are reshaped into a $12 \times 16$ 2D grid.
\end{itemize}
Finally, these low-resolution 2D grids are upsampled back to the original $640 \times 480$ resolution via bicubic interpolation and overlaid onto the original RGB images using a JET colormap.

\subsection{Analysis: Emergent Attention Decoupling}
As illustrated in Fig~\ref{fig:exp-repre}, the visualization reveals a compelling phenomenon: without any explicit pixel-level or grounding supervision, the model naturally learns to decouple semantic understanding from spatial-motor control, perfectly mirroring the ``Ventral-Dorsal'' (What-How) pathways in the human visual system.
The Semantic Stream ($E_{sem}$) acts as the ``What'' pathway. The heatmaps (red boxes) show that $E_{sem}$ tokens are highly activated on task-relevant semantic entities specified in the language instructions. For instance, in the ``Put pomegranate into plate'' and ``Pick up the green pear'' tasks, the attention of $E_{sem}$ sharply focuses on the target fruits and the goal receptacles, while almost completely ignoring the robot's own embodiment. This indicates that $E_{sem}$ is primarily responsible for visual grounding and identifying ``what'' to manipulate.
The Dorsal Stream ($E_{dor}$) acts as the ``How/Where'' pathway. Conversely, the heatmaps (yellow boxes) demonstrate that $E_{dor}$ tokens consistently attend to the robot's end-effector (the gripper), the immediate interaction boundaries, and the spatial relationship between the arm and the objects. In the ``Sweep the medicine box into garbage can'' task, $E_{dor}$ heavily focuses on the broom and the spatial gap between the broom and the box.

\section{Training and Evaluation Details}
\label{sec:app-train}
For experiments on Qwen, we used a learning rate of 1e-5 and employed packed datasets within the FSDP framework to maximize resource utilization. For experiments on Paligemma, we take use of the official implementation of $\pi_{0}$, using a learning rate of 5e-5. For action fine-tuning and evaluation, we adopted different settings for various downstream scenarios:
\begin{itemize}
    \item Calvin ABC-D Simulation Environment: We trained on 8 A800 GPUs (effective batch size 192) for 30k steps. We used an action chunk size of 10. For evaluation, we tested on 1,000 tasks of length 5 from the D-split and reported the mean task completion length.
    \item Robotwin Simulation Environment: We trained on 8 A800 GPUs using 800 clean demonstrations (16 tasks $\times$ 50) for 30k steps. We used an action chunk size of 16, sampling one action every 3 steps (effective action horizon of 48). For evaluation, we tested 100 times on 50 tasks in both Clean settings using unseen instructions and reported the average success rate.
    \item Real-Robot Tasks: We trained on 8 A800 GPUs for 30k steps. We used an action chunk size of 24, inputting three views (primary, left wrist, right wrist). For evaluation, we tested each task type 20 times with randomized initial positions.
\end{itemize}

\section{More Results of Inference Delay}
\label{sec:app-delay}
A potential concern regarding UAM is that its reliance on a larger VLM (7B), coupled with the integration of additional $E_{dor}$ and action experts, may exacerbate the complexity and difficulty of both training and inference. Regarding training, as detailed in Appendix~\ref{sec:app-train} and BagelVLA~\cite{hu2026bagelvla}, straightforward optimizations enable the effective training of UAM even on limited computational resources. As for deployment, it is undeniable that a larger model incurs higher inference latency. To quantify this, we evaluate the single-step inference latency of several different methods under identical GPU hardware and configurations:
\begin{table}[h]
\centering
\vspace{-2ex}
\caption{Inference Speed of Different Models.}\label{tab:app-inf}
\begin{tabular}{lccc}
\toprule
\textbf{Model} & Size & Inference Speed\\
\midrule
$\pi_{0.5}$ & 2.3B + 0.3B MoT & ~250ms\\
Qwen7B + MLP Head & 7B& ~1000ms\\
Qwen7B-$\pi_0$& 7B + 2B MoT& ~1300ms\\
UAM & 7B + 7B + 2B MoT& ~1500ms\\
\bottomrule
\end{tabular}
\end{table}
\vspace{-2ex}

As demonstrated in Tab~\ref{tab:app-inf}, models employing a 7B backbone are notably slower than their 2B counterparts. However, when controlling for the backbone size, incorporating an additional 7B MoT expert exerts a minimal impact on the overall inference speed. Furthermore, UAM adopts a single-step denoising mechanism, bypassing the need for the world model expert to reconstruct the entire image. By directly leveraging the intermediate representations from this initial denoising step for action generation, the 3-expert UAM achieves an inference latency comparable to that of the 2-expert Qwen-$\pi_0$. Ultimately, this marginal increase in inference latency is a well-justified trade-off for the superior generalization performance and enhanced semantic retention capabilities afforded by the additional $E_{dor}$ expert.

\end{document}